\documentclass[10pt,twocolumn,letterpaper]{article}

\usepackage{cvpr}
\usepackage{times}
\usepackage{graphicx, epsfig, epstopdf, pdfpages, wrapfig}
\usepackage{amsmath, amssymb}
\usepackage{color}
\usepackage{subfig}
\usepackage{algorithm,algpseudocode}
\usepackage{bm}
\usepackage{url}
\usepackage{enumitem}
\usepackage{booktabs}
\usepackage{multirow}
\usepackage{verbatim}
\usepackage{mathtools,siunitx}
\usepackage{placeins}

\usepackage[pagebackref=true,breaklinks=true,letterpaper=true,colorlinks,bookmarks=false]{hyperref}

\cvprfinalcopy 

\begin{document}

\title{Exploring Pose Priors for Human Pose Estimation with Joint Angle Representations}

\author{
Yaadhav Raaj (Graduate Student)\\
The Robotics Institute, Carnegie Mellon University\\
{\tt\small \{raaj@cmu.edu\}}
}

\maketitle




\vspace{-.2in}

\section{Introduction}

Pose Priors are critical in human pose estimation, since they are able to enforce constraints that prevent estimated poses from tending to physically impossible positions. Human pose generally consists of up to 22 Joint Angles of various segments, and their respective bone lengths, but the way these various segments interact can affect the validity of a pose. Looking at the Knee-Ankle segment alone, we can observe that clearly, the Knee cannot bend forward beyond it's roughly 90 degree point, amongst various other impossible poses below. 

\begin{figure}[h]
\centering
\includegraphics[width=0.375\textwidth]{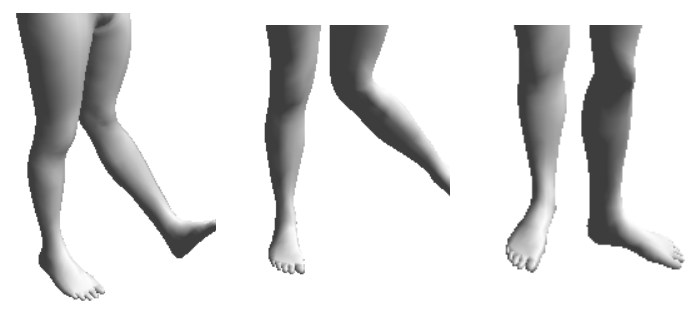}
\vspace{-.1in}
\caption{Various impossible poses of the Knee-Ankle segment as we iterate through all axis of orientations}
\label{fig:teaser}
\end{figure}

However, even within that segment alone, could be various correlations in possible poses, where rotation about one axis closer to a further point can be limit the rotation about another axis. This can also influence the possible range of poses within neighbouring segments, such as the Hip-Knee segment. This can be further influence by bone length as well. Hence, we can see that modelling this can become quite complex. Let us look at several ways in which we can model this, and observe it's flaws. To do this, we fit SMPL \cite{SMPL:2015} to the CMU Mocap Dataset \cite{cmumocap} and get 22 joint angles.

One quick way to model constraints, would be to simply set joint angle limits for each pose and each axis. From Fig. 2, we can already see that even within a unique segment exists various correlations between each axis. This constraint would have worked had the datapoints been axis aligned, but instead are themselves rotated. Furthermore, hard joint limits are not differentiable, and would have to be made soft via some cost function penalizing poses that go out of this topology.

\begin{figure}
\centering
\includegraphics[width=0.475\textwidth]{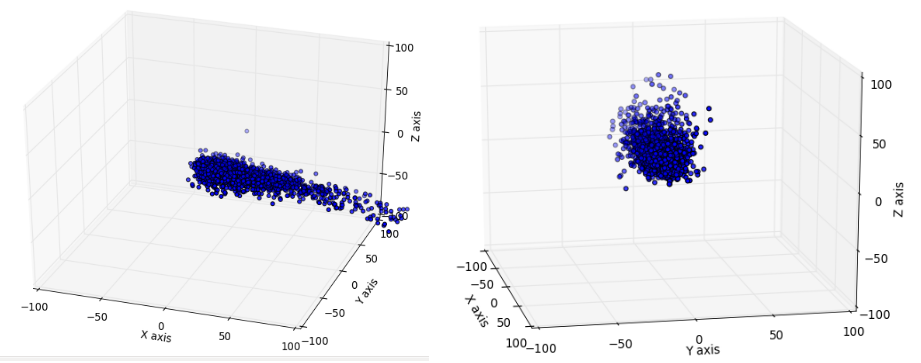}
\vspace{-.1in}
\caption{Plotting the XYZ Joint Angles of the Knee-Ankle and Hip-Knee joints. We observe correlations within a joint segment alone, making simple joint angle limit as a constraint implausible.}
\vspace{-.1in}
\label{fig:teaser}
\end{figure}

This is what initially motivated the Multivariate Normal Distribution Model motivated by \cite{keepitsmpl}, and used by a plethora of papers \cite{keepitsmpl} \cite{vnect} \cite{deephuman} \cite{hmrKanazawa17} \cite{biolstm} \cite{pedx} \cite{totalcapture} in the computer vision pose estimation domain. It elegantly captures the correlation of datapoints by computing the covariance and mean of all 22 * 3 joint angles, and produces a computationally efficient differentiable model whose PDF can be calculated via the function below.

\vspace{-.2in}
\begin{equation}
    f_(x)=\frac{1}{\sqrt{(2\pi)^{n}|\boldsymbol{\Sigma}|}}\exp\left(-\frac{1}{2}({x}-{m})^{T}{\boldsymbol{\Sigma}}^{-1}({x}-{m})\right)
\end{equation}

This model is fit with a large set of datapoints from the CMU Mocap \cite{cmumocap} and Human 3.6M \cite{human3.6m} datasets consisting of over 65k poses. However, many of the above papers mentioned that the model did struggle with poses that were close to the joint limits or over penalized more challenging poses, especially with highly articulated joints such as Knee, Shoulder etc. To verify why this happens, we randomly sample 3000 points from these datasets, and fit just the Wrist and the Knee segments to the above distribution. We then rescale and reorient the datapoints along its principal eigenvectors, and attempt to plot axis of the datapoints with the highest reoriented eigenvalue (magnitude), via the expressions below, applying Singlular Value Decomposition (SVD) on the Covariance of datapoints, and fitting a 1D Normal Distribution on the 1st Principal Component datapoints:

\vspace{-.2in}
\begin{align}
\Sigma_{P}=U\Sigma_{diag}V^{T}
\\
P_{oriented}=U*(P-\mu_{P})
\end{align}

\begin{figure}
\centering
\includegraphics[width=0.475\textwidth]{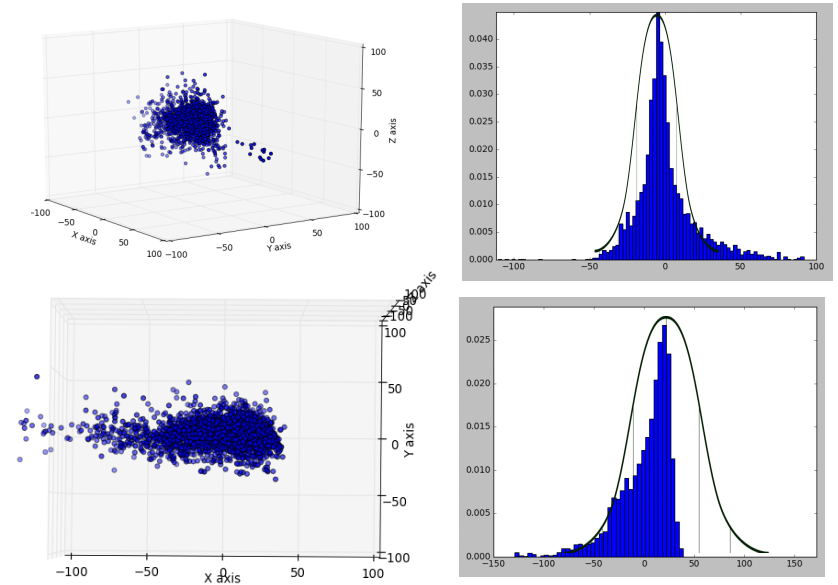}
\vspace{-.2in}
\caption{Attempting to fit a Normal Distribution on the 1st Principal Component of the Wrist Joint Angles and Knee Joint Angles respectively}
\label{fig:teaser}
\end{figure}

We are immediately apple to note two major issues. Let us look at the top of Figure 3, noting the Wrist data. Firstly, by the nature of the dataset, we can observe that the Mean pose obviously tends to an angle of 0. If we attempt to fit a Normal Distribution on this data, inherently poses that are nearing the joint angle limit are going to be less likely in the dataset, and are going to be penalized with a lower probability during test time. What we instead want perhaps is for all possible valid poses to be equally likely, and a Normal distribution is unable to represent this. Furthermore, if you look at the bottom of Figure 3, you can see that with the Knee segment, the most likely pose is obviously at Angle 0, and the knee if able to bend backward up to 90 degrees. However, it cannot possibly bend forward as seen in the data. Hence fitting a normal distribution here would result in impossible poses being given a high probability. 

\begin{gather}
f(x;\alpha;\beta)=\frac{\left(\beta^{alpha}x^{\alpha-1}e^{-\beta x}\right)}{\Upsilon(\alpha)}\\
p(\theta)=\sum_{i=1}^{K}\phi_{i}\mathcal{N}(\mu_{i},\Sigma_{i})
\end{gather}

We could attempt to fit a multivariate gamma distribution as seen in the equation below, but this does not solve the 1st problem of less likely poses getting penalized. Alternatively, we could also try a gaussian mixture model (GMM) as seen above, with $\phi_{i}$ as a control parameter, and this has been attempted in the work by PedX \cite{pedx}. In fact, the work in PedX had actually proposed modelling the temporal prior as a GMM as well. 

\vspace{-.4in}
\begin{gather}
\triangle x^{t}\sim\sum_{i=1}^{K}\phi_{i}\mathcal{N}(\triangle x^{t},\mu_{i},\Sigma_{i})
\end{gather}

This is modelled in such a way where the vector components consist of $\triangle x=(\triangle t,\triangle\theta)$ which captures the angular velocities and in this case the translation of the joint angles, and conditions those in order to predict the likelihood of a pose prediction. This is certainly an interesting direction as given the velocity of a segment, the probability of a particular pose can change conditioned on that angular velocity. If we know that the knee segment is moving in a certain axis, and given the current joint angle, our new joint angle has to be within some limited range. This has been the motivation of Extended Kalman Filter based approaches

\begin{figure}[h]
\centering
\includegraphics[width=0.275\textwidth]{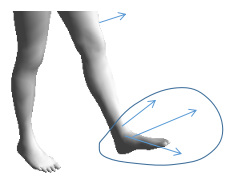}
\caption{Given the current joint angle of the Knee and it's angular velocity, the set of possible pose candidates becomes limited}
\label{fig:teaser}
\end{figure}

The reason we are using such existing statistical distributions is because we need to build a differentiable function that can smoothly separate the impossible and possible poses. Since our dataset only consists of positive valid poses, it is not easy to build a non-linearly separable cost function via Neural Networks.  What we could do however, is to use Biomechanics knowledge and constraints to create a set of poses that are invalid, and attempt to build a binary classifier that predicts if a pose is valid or not. Alternatively, we could use a VAE (Variational Auto Encoder) to build a non-linear transformation function that encourages and warps the final latent space to follow a normally distributed 0 mean centered form. This is what is done in the following work \cite{smplx}.

\vspace{-.2in}
\begin{gather}
Z=Encoder(R)\\
L_{total}=L_{KL}+L_{rec}+L_{orth}+L_{det1}+L_{reg}\\
L_{KL}=KL(q(Z|R)||\mathcal{N}(0,1))\\
L_{rec}=||R-\hat{R}||_{2}^{2}\\
L_{orth}=||\hat{R}\hat{R}'-I||_{2}^{2}\\
L_{det1}=|det(\hat{R})-1|\\
L_{reg}=||\theta||_{2}^{2}
\end{gather}
 
This effectively allows us to warp the distribution of the Knee joint seen in Figure 3 behave more like a normal distribution from it's gamma distribution like shape, albeit in a much more complex multidimensional sense via the VAE. However, it would certainly be interesting to also explore this method over the introduction of negative samples to effectively train a linearly separable space via non-linear transformations.

Overall, we can see that in general, using Neural Networks via an Autoencoder seems to show the most promise as far as representing the complexity and validity of the human pose, and we can see that incorporating and conditioning these priors on newer inputs such as angular velocity and possibily acceleration given the existing pose or state, can certainly be a possible direction going forward.

\FloatBarrier
{\small
\bibliographystyle{ieee}
\bibliography{egbib}
}

\end{document}